\documentclass{article}

\usepackage{amsmath}
\usepackage{PRIMEarxiv}
\usepackage[super]{nth}
\usepackage[utf8]{inputenc} 
\usepackage[T1]{fontenc}    
\usepackage{hyperref}       
\usepackage{url}            
\usepackage{booktabs}       
\usepackage{amsfonts}       
\usepackage{nicefrac}       
\usepackage{microtype}      
\usepackage{lipsum}
\usepackage{fancyhdr}       
\usepackage{graphicx}       
\graphicspath{{media/}}     

\title{KAT to KANs: A Review of Kolmogorov-Arnold Networks and the Neural Leap Forward
}

\author{
  Divesh Basina, Joseph Raj Vishal, Aarya Choudhary,  Bharatesh Chakravarthi\\
  Arizona State University \\
  \texttt{dbasina@asu.edu, jnolas77@asu.edu, achoud65@asu.edu, bshettah@asu.edu} \\
}

\begin{document}
\maketitle

\begin{abstract}

The \textit{curse of dimensionality} poses a significant challenge to modern multilayer perceptron-based architectures, often causing performance stagnation and scalability issues. Addressing this limitation typically requires vast amounts of data. In contrast, Kolmogorov-Arnold Networks have gained attention in the machine learning community for their bold claim of being unaffected by the curse of dimensionality. This paper explores the Kolmogorov-Arnold representation theorem and the mathematical principles underlying Kolmogorov-Arnold Networks, which enable their scalability and high performance in high-dimensional spaces.
We begin with an introduction to foundational concepts necessary to understand Kolmogorov-Arnold Networks, including interpolation methods and Basis-splines, which form their mathematical backbone. This is followed by an overview of perceptron architectures and the Universal approximation theorem, a key principle guiding modern machine learning.  This is followed by an overview of the Kolmogorov-Arnold representation theorem, including its mathematical formulation and implications for overcoming dimensionality challenges.
Next, we review the architecture and error-scaling properties of Kolmogorov-Arnold Networks, demonstrating how these networks achieve true freedom from the curse of dimensionality. Finally, we discuss the practical viability of Kolmogorov-Arnold Networks, highlighting scenarios where their unique capabilities position them to excel in real-world applications. This review aims to offer insights into Kolmogorov-Arnold Networks' potential to redefine scalability and performance in high-dimensional learning tasks.

\end{abstract}


\section{Introduction}

In April $2024$, Liu et al. \cite{liu2024kan} introduced Kolmogorov-Arnold Networks (KANs) as a foundational framework for neural network architecture, attracting attention for its advancements in the field. While previous works have explored applications of the Kolmogorov-Arnold representation theorem (KAT) within deep learning contexts \cite{schmidt2018kolmogorov, selitskiy2022kolmogorov}, not many have applied the theorem with the comprehensive scope achieved by Liu et al. \cite{liu2024kan}. By reinterpreting KAT through a novel architectural lens, their work has not only revitalized interest across the neural network community. Still, it has also opened new avenues for research across diverse application areas. Since introducing KANs, a range of extensions and practical applications have emerged. Researchers are now exploring KANs in varied fields, such as time-series analysis, where the architecture shows promise in capturing complex temporal patterns \cite{vaca2024kolmogorov}. KANs have been used in conjunction with recurrent neural network architecture to develop models that can perform multi-step time series forecasting \cite{genet2024tkan}. In computer vision, early studies suggest that KANs can compete with, and even surpass, traditional architectures like multilayer perceptrons (MLPs) for specific visual processing tasks \cite{cheon2024efficacy, yu2024kan_mlp_comparison}. Additionally, a recent adaptation known as Wav-KAN integrates wavelet transformations with KANs, paving the way for efficient techniques in signal processing \cite{bozorgasl2024wavkan}. Furthermore, KANs have been extensively implemented for scientific discovery in identifying relevant features, modular structures, and symbolic formulas \cite{liu2024kan2,koenig2024kan,howard2024finite}. In quantum physics, KANs were used to design a quantum architecture search model \cite{kundu2024kanqas} due to the significantly fewer parameters it requires compared to MLPs. Moreover, KANs have also been studied in computational biomedicine to generate reliable models for the same reason that they need far less data, relative to MLP-based models \cite{samadi2024smooth}. Together, these applications highlight the versatility and far-reaching potential of KANs within and beyond machine learning.

In response to the growing interest, this article provides a foundational overview of the core concepts needed to understand and implement the  KAT and KANs. It begins with an introduction to the KAT, discussing its origins, mathematical basis, and specific advantages in neural network design demonstrating why KANs offer a valuable alternative to traditional models in machine learning. The article then provides a key mathematical background for KANs, focusing on interpolation techniques, especially B-splines, which enable efficient function approximation within KAN architectures. To further contextualize KANs, we revisit the perceptron model and universal approximation theorem (UAT). This review of foundational concepts highlights the limitations of conventional architectures and underscores the relevance of KANs as a complementary approach. Building on this foundation, the article explores the structure of KANs, describing how KAT integrates into a neural network architecture. It also examines the error-scaling characteristics that set KANs apart, showcasing their unique computational efficiency and approximation accuracy strengths. Finally, the article discusses the practical implications of KANs, particularly in terms of scalability and error behavior, and their potential to drive advancements in neural network theory. By situating KANs within the broader context of machine learning architectures, this review emphasizes their distinctive contributions and the opportunities they present for further research and applications.

\section{Background}

In $1900$, at a summit in Paris, David Hilbert presented $23$ problems that would become foundational to modern mathematics \cite{hilbert1984mathematical}. Among these was the $\nth{13}$ problem, which dealt with the representation of polynomial equations and posed profound questions about the limitations of expressing solutions to these equations using elementary functions and operations \cite{hilbert1927mathematische}. By the early $\nth{20}$ century, mathematicians had developed methods to solve polynomial equations up to the fourth order using radicals and basic arithmetic operations \cite{wikipedia_kolmogorov_arnold}. However, the pioneering work of Evariste Galois revealed that, beyond the fourth order, solutions to polynomial equations could not generally be expressed in such elementary terms \cite{galois1962theory}.

To address these challenges, transformations like the Tschirnhaus transformation were introduced. This transformation could reduce an algebraic equation of order 
\textit{n} written in the form:
\[x^n + a_{n-1}x^{n-1} + ... a_0 = 0 \] to a simplified form: \begin{equation}\label{1}
y^n + b_{n-4}y^{n-4} + ... b_1y + 1 = 0 
\end{equation}
This reformulation allowed certain higher-order polynomial equations to be represented as a superposition (a combination) of functions of fewer variables, specifically as functions of two variables when $n<7$ and $n-4$ variables when 
$n\geq7$. For instance, it was believed that an equation of order $n=7$, \begin{equation} y^7 + b_3y^3 + \dots + b_1y + 1 = 0 \end{equation} could not be further reduced into a representation using only lower-order functions of fewer variables.

Hilbert concluded that a general simplification of a seventh-order polynomial equation into a superposition of lower-order functions was likely unattainable. It was in response to this conjecture that Andrey Kolmogorov and Vladimir Arnold made a revolutionary contribution to the field. They demonstrated that any continuous multivariate function defined from \([0,1]^n\rightarrow \mathbb{R}\) could be represented as a superposition of continuous uni-variate functions \cite{kolmogorov1957representation, arnold1958representation}. This theorem, known as the Kolmogorov-Arnold representation theorem, presented a breakthrough in understanding function complexity, showing that even seemingly intricate multivariate functions could be decomposed into a manageable combination of simpler, single-variable functions. Kolmogorov and Arnold's work laid a mathematical foundation that would later prove highly relevant to modeling complex behaviors in various fields, including machine learning and neural networks. In recent times, the theorem has found new significance in the form of KANs, which apply KAT to enhance machine learning applications. KANs are now being actively explored for tasks such as object recognition \cite{azam2024suitability}, natural language processing \cite{yu2024kan_mlp_comparison}, and time series prediction \cite{hou2024comprehensive_kan, vaca2024kolmogorov}, where they provide new avenues for efficient and interpretable model architectures.

\section{Mathematical Foundation}
\label{sec:headings}

KAT \cite{kolmogorov1957representation, arnold1958representation} asserts that any multivariate continuous function \( f: [0,1]^n \rightarrow \mathbb{R}\) can be represented as a superposition of continuous single-variable functions. Specifically, it states:
\[f(x) = f(x_1,x_2,...,x_n) = \sum_{q=0}^{2n} \Phi_q  \left(\sum_{p=1}^{n} \phi_{q,p}(x_p)\right) \]
where each $\phi_{q,p}$, is defined as, \[\phi_{q,p}: [0,1] \rightarrow \mathbb{R},\] and \[\Phi_q: \mathbb{R} \rightarrow \mathbb{R}.\] This decomposition enables the approximation of high-dimensional functions through simpler univariate components.

In the context of KANs, under the assumption that the process we are trying to model has a higher-order polynomial expression, we can leverage Galois theory and KAT to numerically approximate \(f(x)\) as a composition of uni-variate functions \cite{galois1962theory,liu2024kan}. The general idea for the implementation of KANs is to approximate each \(\phi_{q,p}\) through B-spline interpolation and sum up the results which serve as the input parameter for \(\Phi_q\) \cite{liu2024kan}.  
This section establishes the necessary mathematical foundations for understanding how KANs work. It begins by exploring the concepts of splines and interpolation, before focusing on B-spline interpolation, which plays a crucial role in the construction of KANs.

\subsection{Interpolation}

\begin{figure}[h]
\centering
\includegraphics[width=0.60\linewidth]{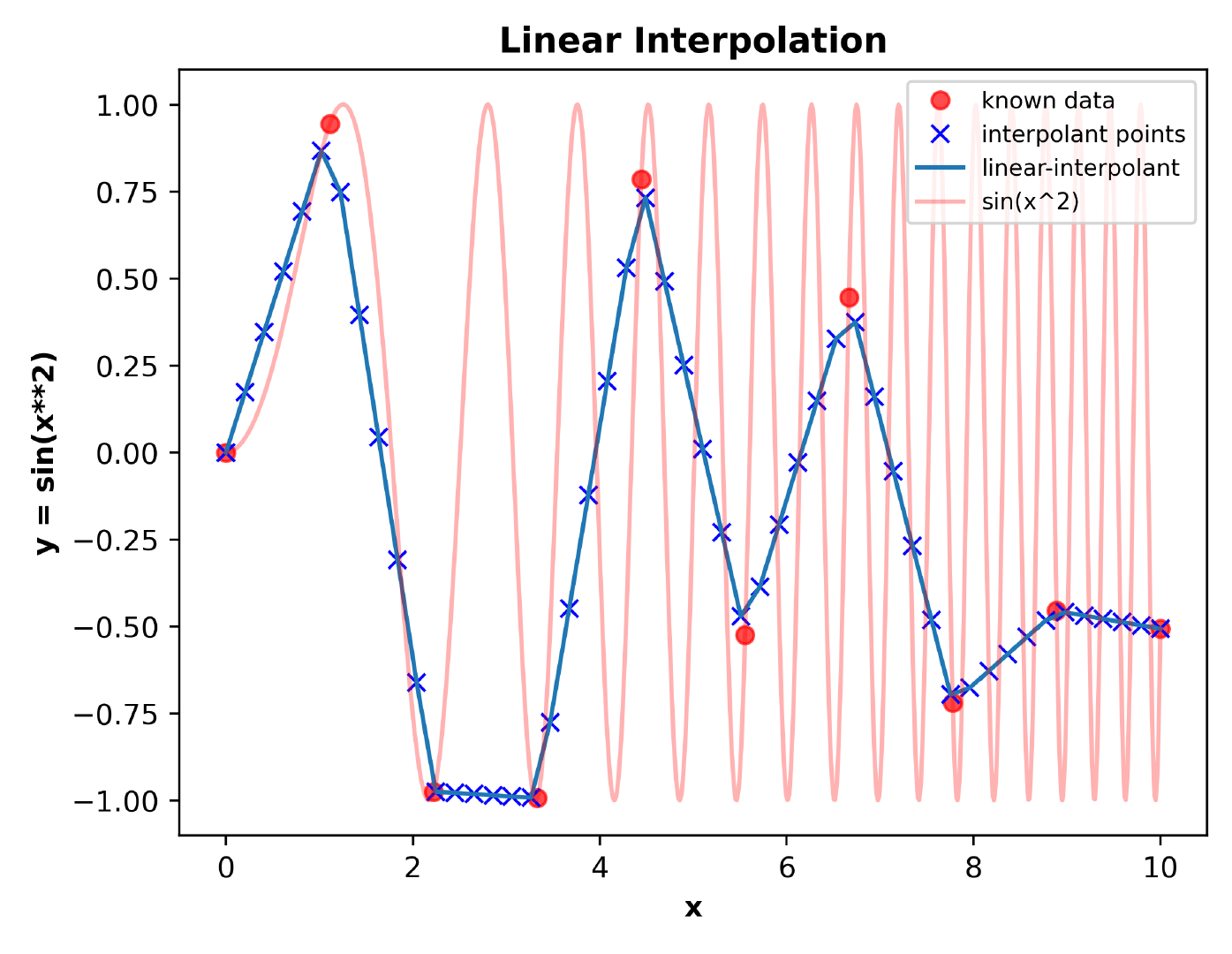}
\caption{Linear interpolation with $50$ interpolant points - the linear interpolant struggles to capture the curvature of the function, resulting in a piecewise linear approximation. While linear interpolation is computationally efficient and provides a smooth curve with sufficient data points, it may not be ideal for accurately modeling functions with significant curvature.}
\label{LinearInterpolation}
\end{figure}

Interpolation is a technique used to estimate unknown data utilizing a given set of discrete data points. The goal is to find a mathematical function that closely represents the provided data, allowing us to predict values at input points where data is not available \cite{burden2010numerical}. Various interpolation methods, such as linear, polynomial, Hermite, and spline interpolation, can be applied depending on the nature of the data \cite{wikipedia_interpolation, burden2010numerical}.

For example, given two data points \((x_a,y_a)\) and (\(x_b,y_b)\),  linear interpolation can be derived using the 2-point form as follows:
\begin{equation}
 \frac {y - y_a}{y_b - y_a} = \frac {x - x_a}{x_b - x_a} 
\end{equation}
Rearranging this, we get the equation for the linear interpolant:
\begin{equation}
y = y_a + (y_b - y_a) \frac{x - x_a}{x_b - x_a}
\end{equation}

The idea behind this method is that for any point  \((x,y)\) between \((x_a,y_a)\) and (\(x_b,y_b)\), the slopes of  lines connecting points \((x_a,y_a)\),\((x,y)\) and \((x,y)\),\((x_b,y_b)\) must be equal. Figure \ref{LinearInterpolation} shows an interpolation example with $50$ interpolant points. 

Polynomial interpolation generalizes the concept to more than two data points. 
Given a set of \(n+1\) bi-variate data points \((x_0,y_0)\),...,\((x_n,y_n)\) we can determine a unique polynomial \(p(x)\) that passes through all the points \cite{wikipedia_polynomial_interpolation}. To do so, we first define the Lagrange Basis polynomial as:
\begin{equation}
    L_i(x) = \prod_{{j \neq i},{j = 0}}^{n} \frac{x - x_j}{x_i - x_j}
\end{equation}

Then, using these basis functions, the interpolating polynomial 
$p(x)$ is expressed as:
\begin{equation}
    p(x) = \sum_{i = 0}^ n L_i(x).y_i
\end{equation}
This method provides a smooth polynomial curve that passes through all the data points, making it useful for fitting complex datasets.

\subsubsection{Splines}
\begin{figure}[h]
    \centering
    \includegraphics[width=0.60\linewidth]{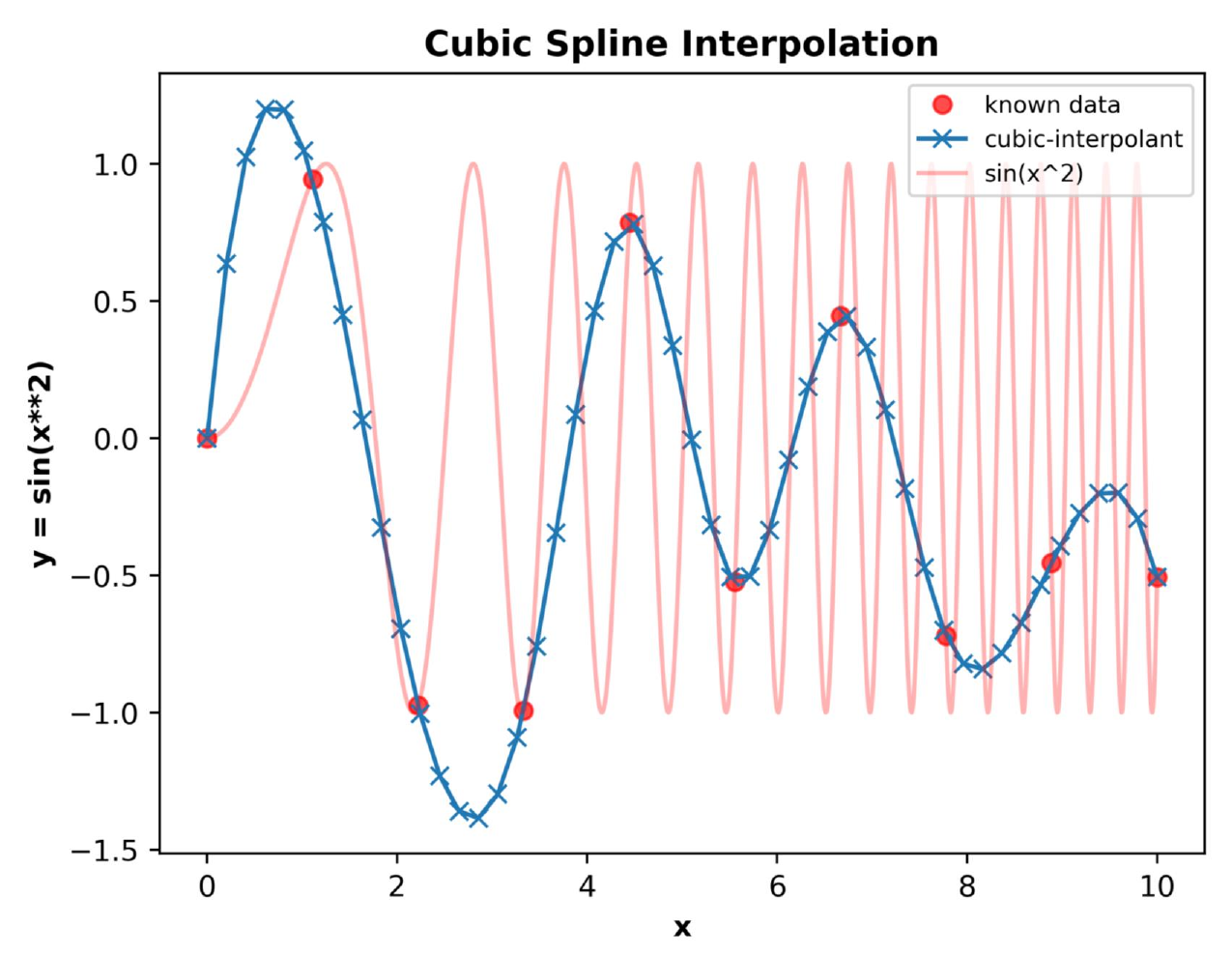}
    \caption{Cubic spline interpolation with $50$ interpolant points - The cubic spline interpolation notably improves the ability to capture the function's curvature compared to linear interpolation, offering a smoother and more accurate fit. Figure \ref{LinearVsCubic} further demonstrates how B-spline interpolation provides enhanced control, improving the precision and smoothness of the interpolant.}
    \label{cubicSplineInterpolation}
\end{figure}

Spline interpolation is used to approximate functions by fitting piece-wise polynomials, rather than using a single high-degree polynomial.
This approach helps to avoid Runge's phenomenon, wherein a high-degree polynomial interpolation can result in excessive oscillation between points \cite{wikipedia_interpolation,wikipedia_spline_interpolation}. Among the various types of spline interpolation, cubic splines are preferred over linear or quadratic splines due to their smooth behavior. Specifically, cubic splines provide continuous first and second-order derivatives, which are generally desired for smooth transitions between data points.

Given a set of data points or knots \((x_0,y_0), (x_1,y_1),....,(x_n,y_n)\),  the goal is to find a set of \textit{n} piecewise cubic equations \(q_i(x)\) for each interval between adjacent knots. These cubic functions must pass through 
the given points and satisfy certain boundary conditions, including matching the first derivatives (slopes) at the endpoints.
Formally, for each segment between the points \((x_{i-1} , y_{i-1})\) and  \((x_i, y_i)\),   the following conditions must hold \cite{wikipedia_spline_interpolation}:

\begin{equation}
    q_i(x_{i-1}) = y_{i-1}
\end{equation}
\begin{equation}
    q_i(x_{i}) = y_{i}
\end{equation}
\begin{equation}
    q'_i(x_{i-1}) = k_1
\end{equation}
\begin{equation}
    q'_i(x_i) = k_2
\end{equation}

Here, $q_i(x)$
represents the cubic polynomial for the interval \(i\), and 
$k_1$ and $k_2$ are the slopes at the endpoints  
$x_{i-1}$ and $x_i$ respectively.
The cubic interpolant is then defined as:
\begin{equation}
    q_i(x) = (1 - t(x)) y_{i-1} + t(x) y_i + t(x)(1-t(x)) \left((1-t(x)) a + t(x) b\right) \\
\end{equation}

Where $t(x)$ is a function defined as:
\begin{equation}
    t(x) = \frac{x-x_1}{x_2 - x_1}
\end{equation}
Additionally, the coefficients 
$a$ and $b$ are given by:
\begin{equation}
    a = k_1(x_2 - x_1) - (y_2 - y_1)
\end{equation}
\begin{equation}
    b = -k_2(x_2 - x_1) + (y_2 - y_1)
\end{equation}
To fully determine the cubic spline, as illustrated in Figure \ref{cubicSplineInterpolation}, 
we solve for coefficients using the boundary conditions on \(q_i, q_{i+1}\) and \(q'_i,q'_{i+1}\) and \(q''_i,q''_{i+1}\) being the same for each \((x_i, y_i)\). These conditions involve ensuring that both the first and second derivatives of the adjacent cubic polynomials match at the internal knots. Depending on the type of spline being used (e.g., natural, clamped, or not-a-knot), additional constraints are applied to the boundary conditions. This approach provides a piecewise smooth function that minimizes oscillations and ensures a better fit to the data, especially when dealing with higher-order interpolations.

\subsection{B-Splines}

\begin{figure}[h]
    \centering
\includegraphics[width=0.91\linewidth]{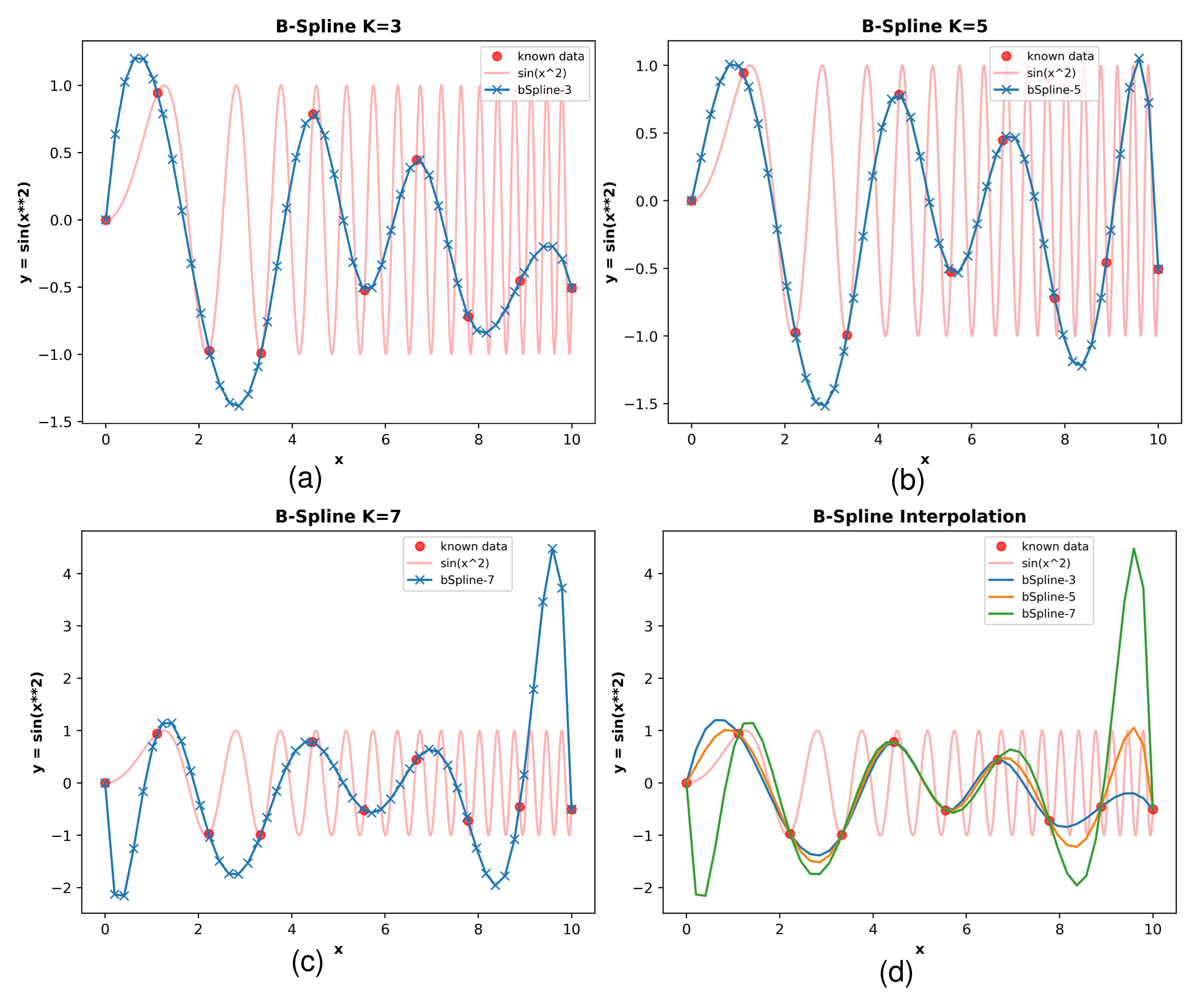}
    \caption{B-spline interpolation of order $k=3$ with $50$ interpolant points, using basis splines to approximate a cubic spline. The figure illustrates a basis spline with $k=5$.}
    \label{B-Spline_k=3_5_7}
\end{figure}

\begin{figure}[h]
    \centering
    \includegraphics[width=0.75\linewidth]{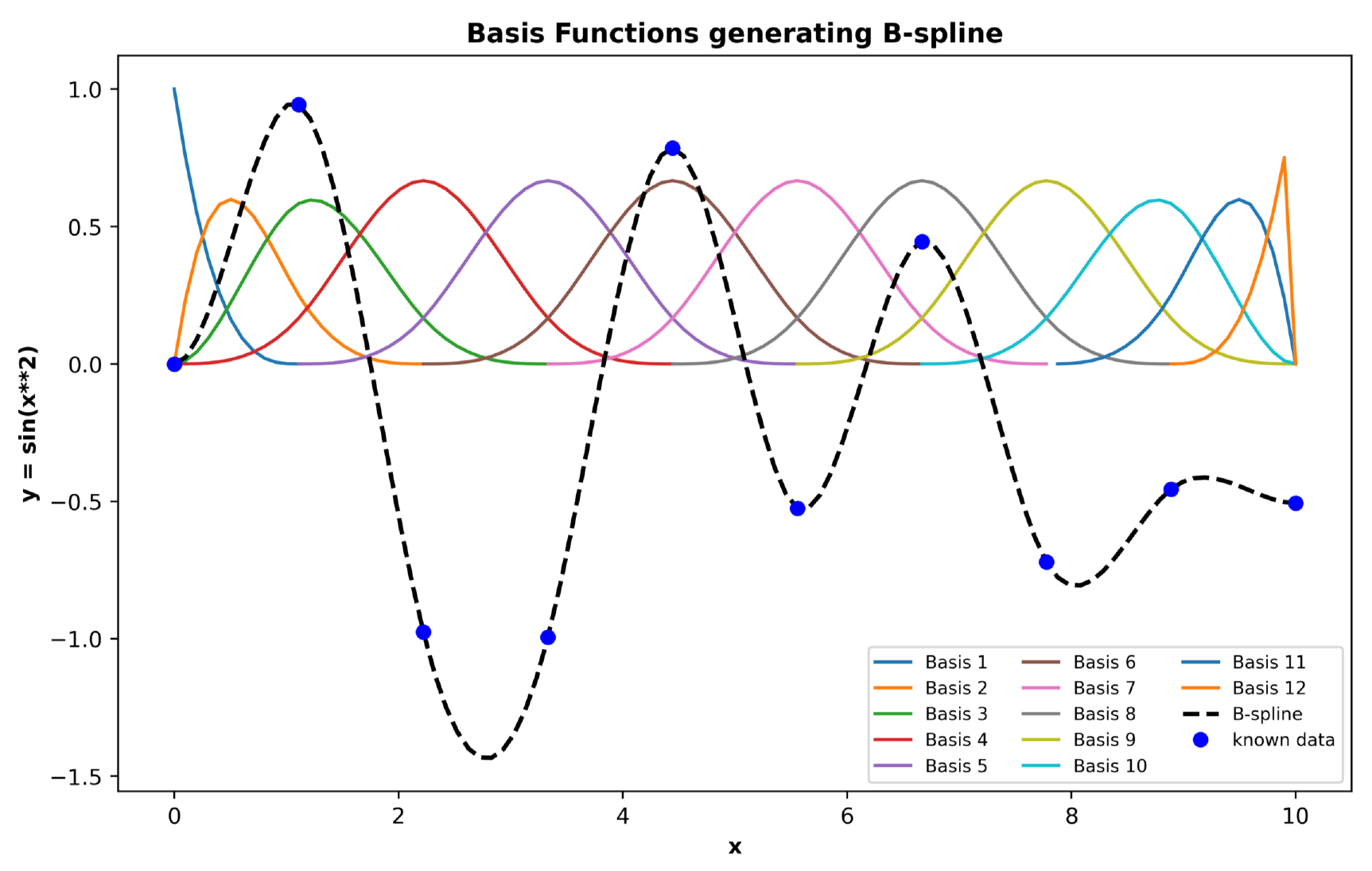}
    \caption{Basis functions used to perform the B-spline interpolation with $k=3$ shown above. This image shows how various basis functions influence the target function generating a smooth curve. For $k=3$ our basis functions have their influence on the target up to $4-knot$ intervals. In general, they exert their influence over $k+1$ knots \cite{wikipedia_bspline}.}
    \label{basisFunction_B-Spline}
\end{figure}

\begin{figure}
    \centering
    \includegraphics[width=0.60\linewidth]{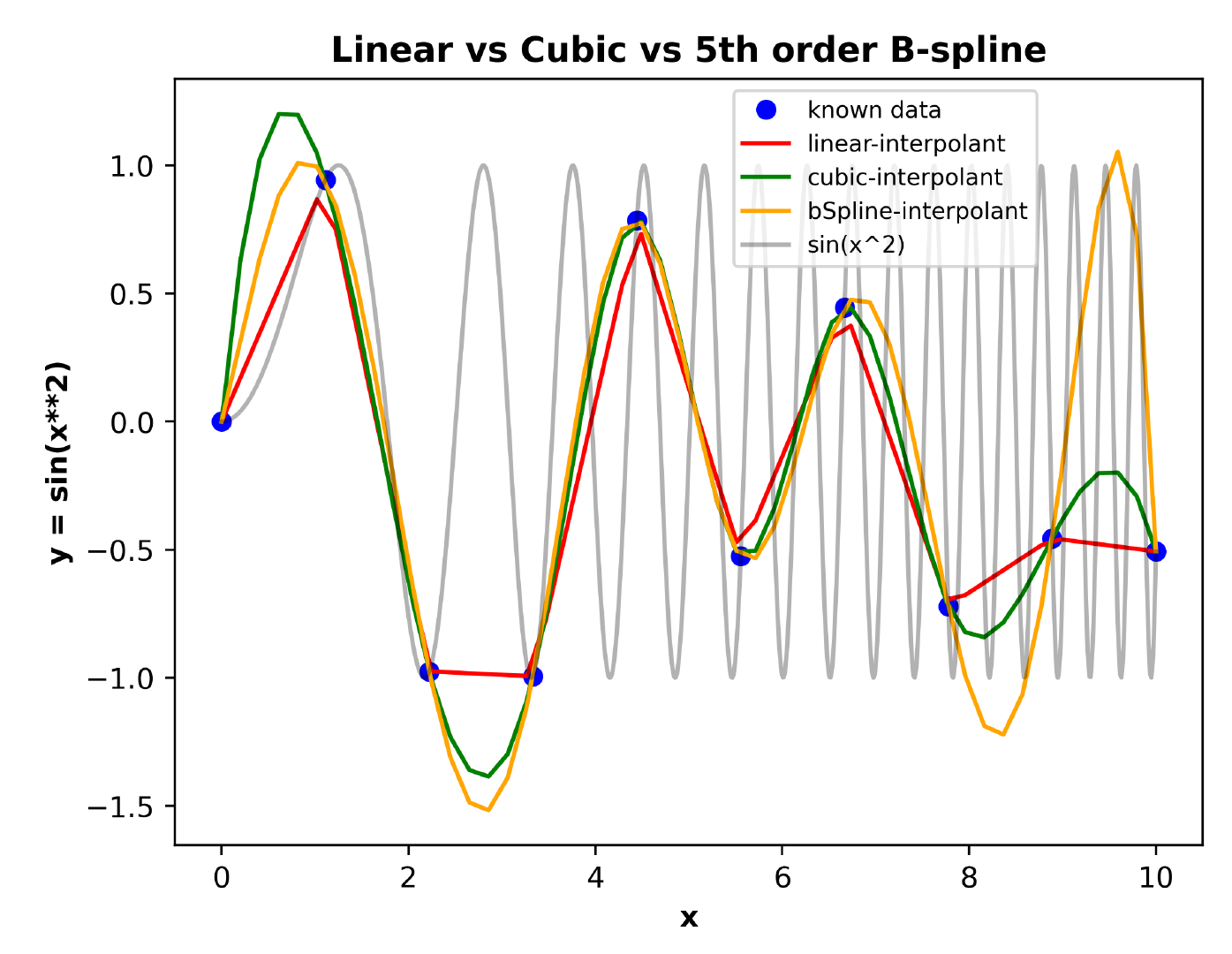}
    \caption{Different Types of interpolation performed with $50$ interpolant points juxtaposed on each other. We used a \(5^{th}\) order ($k=5$) B-spline to show the higher-order capabilities of using b-spline functions over cubic splines. As we can see here, in the interval [8,10], the b-spline interpolant manages to interpolate the extreme points of the function without needing any data.}
    \label{LinearVsCubic}
\end{figure}

B-splines of order \textit{$n$} are basis functions for splines of the same order, where, any spline can be represented as a linear combination of B-splines. Additionally, there exists a unique combination of B-splines for any given spline function \cite{wikipedia_bspline, deboor2001splines}.
To define these basis functions, we consider a set of non-decreasing knots \(\textbf{t} = (t_0,t_1...t_m)\) and construct a basis polynomial \(B_{i,p}(t)\) of degree \(p\) and order \(p+1\) such that the polynomial is defined as:
\begin{equation}
        B_{i,p}(t) =
    \begin{cases}
        non-zero,  &t_i \leq t \leq t_{i+p+1} \\
        0, & \text{otherwise}
    \end{cases}
\end{equation}
With the added constraint:
\begin{equation}
    \sum_{i=0}^{m-p-1} B_{i,p}(t) = 1
\end{equation}
which ensures that the scaling factor of \(B_{i,p}(t)\) is fixed across the interval. We define \(B_{i,p}(t)\) recursively, starting with \(B_{i,0}(t)\) and progressively build up to higher orders \cite{wikipedia_bspline}. The zeroth-order basis function is then defined as: 
\begin{equation}
    B_{i,0}(t) = 
    \begin{cases}
        1, & t_i \leq t \leq t_{i+1} \\
        0, & otherwise
    \end{cases}
\end{equation}
For higher orders, we recursively define $B_{i,p}(t)$ as: 
\begin{equation}
    B_{i,p}(t) = \frac{t-t_i}{t_{i+p} - t_i}B_{i,p-1}(t) + \frac{t_{i+p+1} - t}{t_{i+p+1} - t_{i+1}}B_{i+1,p-1}(t)
\end{equation}
With basis functions, we define a spline of order \(p\) over a given knot vector \textbf{t} as \cite{wikipedia_bspline}: 
\begin{equation}
    S_{n,\textbf{t}}(x) = \sum_i \alpha_i B_{i,p}(x)
\end{equation}
where, $\alpha_i$ are the coefficients of the spline and $B_{i,p}(x)$ are the corresponding B-spline basis functions. Figure \ref{B-Spline_k=3_5_7} illustrates the veracity of B-splines of different orders, whereas Figure \ref{basisFunction_B-Spline} displays the effect of various basis functions over a spline interpolant. Figure \ref{LinearVsCubic} contrasts the differences among the various interpolation techniques.

\subsection{Perceptrons}
Perceptrons serve as the building blocks of neural networks. A perceptron accepts features \({x_1,x_2,x_3,...,x_n}\), weights \({w_1,w_2,w_3,..,w_n}\) and  bias term \(b\) as inputs to the perceptron function \(\phi(x,w)\).  The perceptron function \(\phi: [0,1] \rightarrow \mathbb{R}\) is mathematically defined as:
\begin{equation}
    \phi(x,w) = \ g(\sum_{i=1}^{n} x_i w_i + b)
\end{equation}

\begin{figure}[h]
    \centering
\includegraphics[width=0.40\linewidth]{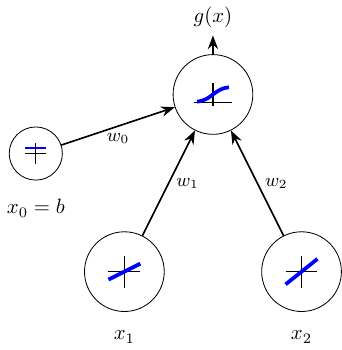}
    \caption{Perceptron with bias $x_{0}$ and inputs $x_{1}$ and $x_{2}$ along with their corresponding weights $w_0$, $w_1$ and $w_2$. Although $w_0$ has been shown explicitly here, often it is generalized as \(b\).}
    \label{perceptron}
\end{figure}

Where \(\ g(x)\), represents the activation function, is commonly the hyperbolic tangent \(tanh\) or the the rectified linear unit $(ReLU)$ activation function \cite{goodfellow2016deep, wikipedia_perceptron}. Figure \ref{perceptron} illustrates the organization of a perceptron with 2 inputs and a bias term. 

\section{Foundational Theories in Neural Model Representation}

Modern machine learning models build upon the perceptron architecture, expanding it into vast networks with multiple, well-organized layers designed to address complex tasks. Each layer is finely tuned to optimize information flow, enhancing the model’s ability to capture intricate relationships and patterns.
A key theoretical foundation that supports these architectures is the UAT which establishes that a neural network with a sufficient number of layers and appropriate structure can approximate a wide range of continuous functions. It assures that neural networks are capable of representing complex functions, though it does not guarantee convergence or optimality of the solution. Rather, it confirms the potential for networks to achieve accurate function approximations, serving as the theoretical groundwork for a diverse array of architectures.

Building on these principles, specialized neural network architectures such as convolutional neural networks (for image processing) and recurrent neural networks (for sequential data) illustrate the UAT’s broad applicability across different domains. In addition, the KAT further extends our understanding of function approximation by showing that any multivariate continuous function can be expressed as a superposition of single-variable continuous functions. This alternative perspective inspires unique architectural designs, like KANs, that leverage function decomposition for enhanced modeling efficiency and precision.
Together, UAT and KAT establish the theoretical feasibility and flexibility of neural networks, guiding the development of modern architectures that address a wide range of machine-learning tasks.

\subsection{Universal Approximation Theorem}
The Universal approximation theorem \cite{wikipedia_universal_approximation_theorem, cybenko1989approximation} states that within a given function space, for every function \(f\) in that space, there exists a sequence of artificial neural networks \(\phi_1, \phi_2, \phi_3...\) such that \(\phi_n \rightarrow f\) as \(n \rightarrow \infty\). 
Additionally, the theorem guarantees that for any desired error bound \(\epsilon > 0\), a sufficiently large neural network can approximate \(f\) within this margin of error \(\epsilon\). However, it neither specifies how to construct or find this sequence of neural networks nor does it provide an explicit method to train such networks to achieve the desired approximation. Moreover, there is no defined bound on the number of neurons \(N(\epsilon)\) required to achieve the error tolerance \(\epsilon\)  \cite{cybenko1989approximation}. 

In addition to the original formulation, there are variations of the theorem that apply to networks with different configurations, such as arbitrary width, arbitrary depth, or bounded widths and depths. These variants ensure that the approximation error remains within the specified tolerance 
\(\epsilon\)  for a given \(\epsilon > 0\) , although with varying levels of efficiency and complexity. These insights contribute to the design of neural networks that are optimized for both the accuracy and computational efficiency required for specific tasks \cite{wikipedia_universal_approximation_theorem}

\subsection{Kolmogorov Arnold Representation Theorem}

The Kolmogorov-Arnold representation theorem, introduced by Vladimir Arnold and Andrey Kolmogorov, states that if \(f\) is a continuous and bounded multivariate function, it can be expressed as the sum of a finite composition of univariate functions \cite{kolmogorov1957representation,arnold1958representation}.  Specifically, for a continuous function \(f: [0,1] \rightarrow \mathbb{R}\), the theorem states that:

\begin{equation}
f(x) = f(x_1,x_2,...,x_n) = \sum_{q=1}^{2n+1} \Phi_q  \left(\sum_{p=1}^{n} \phi_{q,p}(x_p)\right) 
\label{eq:kan_representation}
\end{equation}

where \[\phi_{q,p}: [0,1] \rightarrow \mathbb{R}\] and \[\Phi_q: \mathbb{R}\rightarrow \mathbb{R}\] are continuous functions.

However, this representation does not guarantee the smoothness or continuity of the functions \(\phi_{q,p} (x_p)\). One of the key challenges with this representation in the context of artificial intelligence is that \(\phi_{q,p} (x_p)\)  may be discontinuous or even exhibit fractal behavior, which makes it difficult to learn using traditional methods \cite{liu2024kan}. Despite this, recent work of Liu et al. \cite{liu2024kan} has shown that by extending the Kolmogorov-Arnold representation beyond its original bounds of \([n,2n+1,1]\) where, \(n\) represents the number of inputs, to arbitrary depths and widths, it can still be made practical for machine learning tasks. Such extended representations have been shown to achieve performance comparable to or even exceeding that of models derived from the UAT in certain scenarios\cite{liu2024kan}. This insight opens up new avenues for using KANs in artificial intelligence, enabling more flexible and powerful approximations of complex functions.

\section{Structure of KANs}

Given the KAT representation, Liu et al. \cite{liu2024kan} show that for supervised learning tasks with data having input-output pairs \((x_i,y_i)\), the goal is to approximate the function \(f(x_i)\approx y_i\). This task can be achieved by approximating the uni-variate function \(\phi_{q,p}\) and \(\Phi_q\), as outlined in the KAT representation \cite{liu2024kan}.  Specifically, equation \ref{eq:kan_representation}  guarantees the existence of such functions. In this framework, a KAN with $2n+1$ hidden layers serves as the model for this approximation. While the original KAT does not provide a generalized version for all sizes, Liu et al. \cite{liu2024kan} work extends the KAT representation to include networks with deeper and wider layers. This extension demonstrates that KANs can be effectively applied to modern deep learning tasks, enabling the approximation of complex functions in practical scenarios \cite{liu2024kan}. 

\subsection{Mathematical notation for KAN}
A KAN layer comprises an \(n_{in}\) dimensional input and \(n_{out}\) dimensional output layer which can be represented as a matrix of 1-d functions \cite{liu2024kan}:
\begin{equation}
    \Phi = \{\phi_{q,p}\}\quad
    p = 1,2,...,n_{in} \quad
    q = 1,2,...,n_{out}
\end{equation}

For example, in KAT, for the inner layer we have \(n_{in} = x_1,x_2..x_n\) and \(n_{out} = 1,2,...2n+1 \) while for the outer layer we have \(n_{in} = 1,2,...2n+1\) and \(n_{out} = 1\) \cite{liu2024kan}. \\

A KAN structure with \(L\) layers can be represented as: 
\begin{equation}
    [n_0,n_1....n_L]
\end{equation}
where \(n_i\) represents the number of nodes in the \(i^{th}\) layer. Then, from our previous example KAT gives us a \([n, 2n+1, 1]\) network \cite{liu2024kan}.\\

\begin{figure}[h]
    \centering
    \includegraphics[width=0.45\linewidth]{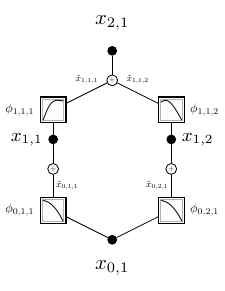}
    \caption{A simple [1,2,1] KAN layer illustrating the above mathematical representation. }
    \label{simpleKAN}
\end{figure}

The \(i^{th}\) neuron of the \(l^{th}\) layer is represented as \((l,i)\). And we have a total of \(n_l*n_{l+1}\) number of activations between consecutive layers \(l\) and \(l+1\). The activation function connecting neurons \((l,i)\) and \((l+1,j)\) is represented as:
\begin{equation}
    \phi_{(l,j,i)},\quad where\quad
    l = 0,1,...L-1\quad
    i = 1,2...n_l\quad
    j = 1,2...n_{l+1}
\end{equation}
The input aka pre-activation for \(\phi_{(l,j,i)}\) is the node \(x_{l,i}\) and the output or post-activation is the edge representing \(\phi_{l,j,i}(x_{l,i}) \equiv \Tilde{x}_{l,j,i}\). The neuron \(x_{l+1, j}\) is then simply a summation of all the post-activations from the previous layer directed to \((l+1,j)\) :
\begin{equation}
    x_{l+1,j} = \sum_{i=1}^{n_l} \phi_{l,j,i}(x_{l,i}) = \sum_{i=1}^{n_l} \Tilde{x}_{l,j,i} 
\end{equation}
In matrix form, we can then represent layer \(x_{l+1}\) as:
\begin{equation}
    \bold{x_{l+1}} = \underbrace{\begin{Bmatrix}
                        \phi_{l,1,1}(.)&\phi_{l,1,2}(.)&...&\phi_{l,1,n_l}(.)\\
                        \phi_{l,2,1}(.)&\phi_{l,2,2}(.)&...&\phi_{l,2,n_l}(.)\\
                        .           &.               &&.             \\
                        .           &.               &&.             \\
                        .           &.               &&.             \\
                        \phi_{l,n_{l+1},1}(.)&\phi_{l,n_{l+1},2}(.)&...&\phi_{l,n_{l+1},n_l}(.)
                          \end{Bmatrix}}_{\Phi_l}\bold{x_l} = 
                        \begin{Bmatrix}
                        \phi_{l,1,1}(x_{l,1})+...+\phi_{l,1,n_l}(x_{l,n_l})\\
                        \phi_{l,2,1}(x_{l,1})+...+\phi_{l,2,n_l}(x_{l,n_l})\\
                        .                                   \\
                        .                                    \\
                        .                                    \\
                        \phi_{l,n_{l+1},1}(x_{l,1})+...+\phi_{l,n_{l+1},n_l}(x_{l,n_l})\\
                          \end{Bmatrix}
\end{equation}
\begin{equation}
    \bold{x_{l+1}} =  
                        \begin{Bmatrix}
                        \Tilde{x}_{l,1,1}+\Tilde{x}_{l,1,2}+...+\Tilde{x}_{l,1,n_l}\\
                        \Tilde{x}_{l,2,1}+\Tilde{x}_{l,2,2}+...+\Tilde{x}_{l,2,n_l}\\
                        .                                   \\
                        .                                    \\
                        .                                    \\
                        \Tilde{x}_{l,n_{l+1},1}+\Tilde{x}_{l,n_{l+1},2}+...+\Tilde{x}_{l,n_{l+1},n_l}\\
                          \end{Bmatrix} 
\end{equation}
Where, \(\Phi_l\) in Eq. (26) is the function matrix of layer \(l\) \cite{liu2024kan}. A general KAN network, illustrated in Figure \ref{simpleKAN}, can then be written as a composition of \(L\) layers as:
\begin{equation}
    KAN(\bold{x}) = (\Phi_{L-1} \circ \Phi_{L-2} ...\Phi_{0})\bold{x}
\end{equation}

\section{KAN's Error and Scaling}

Liu et al. \cite{liu2024kan} theorizes the upper bound on error, proving that the error bound is not reliant on the dimensions of the input \cite{liu2024kan}, leading to an extremely strong claim stating that KAN’s are not subject to the curse of dimensionality \cite{liu2024kan}. The theorem states:
\newtheorem{theorem}{Theorem}
\begin{theorem}
Let $x = (x_1,x_2,x_3...x_n)$. Suppose that a function f(x) admits a representation \[f(x) = (\Phi_{L-1} \circ \Phi_{L-2} ...\Phi_{0})\bold{x}\] as Eq (28), where each $\Phi_{l,i,j}$ are (k+1) - times continuously differentiable. Then there exists a constant C depending on f and its representation,  such that we have the following
approximation bound in terms of the grid size G: there exist k-th order B-spline functions $\Phi_{l,i,j}^G$ such that for any $0\leq m \leq k$, we have the bound \[|| f - (\Phi_{L-1} \circ \Phi_{L-2} ...\Phi_{0})\bold{x}\ ||_{C^{m}} \leq CG^{-k-1+m}\]
\end{theorem}
According to Theorem 1 \cite{liu2024kan}, the error is bounded by a scaled value of the number of grid points used for the basis splines. A fundamental question to consider is what happens when a model is trained with unrelated input and output features. Does the KAN approximation error remain bounded when trained on arbitrary and unrelated inputs and outputs? 

Traditional machine learning algorithms are reliant upon being trained with high dimensional input vectors to minimize error. The problem with this approach is that when adding a new dimension, the feature space to search increases exponentially. Consider this as an example. In the Euclidean space, suppose there are $5$ discrete points on the X-axis. The probability of randomly choosing one of these points is \(\frac{1}{5}\). Suppose we add $5$ more points in the $Y$ direction, representing the addition of a new feature. The probability space transforms into a 2-dimensional \(5\times5\) grid. Given this new space, existing probabilities of the points in $X$, now have probabilities each of \(\frac{1}{25}\). Furthermore, with the addition of each new dimension, it is necessary to provide MLP models with sufficient data to be able to generalize meaningfully. This example describes how the data requirement scales by adding new features/dimensions, also known as the \textit{curse of dimensionality}.

By being able to avoid the curse of dimensionality, KANs allow the training of large and feature-rich models without needing vast input data to achieve high accuracy. This is possible because of the decoupling of error from the input size as shown in Theorem $1$.
 KANs have also been shown to require far fewer parameters due to their intrinsic nature of approximating target functions directly using b-splines, while still maintaining high accuracy. This is made possible by controlling the grid size \(G\) to generate higher resolution B-spline approximations. 

The limitations of the universal approximation theorem are realized in this context. It requires that we have \(k \geq N(\epsilon)\) number of neurons to reach an error tolerance of \(\epsilon\) but we do not have a bound on how \(N(\epsilon)\) scales \cite{wikipedia_universal_approximation_theorem,cybenko1989approximation}. Neural scaling laws state that \(l \propto N^{-\alpha}\)  where, \(l\) is RMSE \cite{sharma2020neural_scaling}. In MLPs, it has been shown \cite{barron1993approximation} that larger \(\alpha\) improved performance, and therefore, scaling was the straightforward way to improve performance. KANS offers a solution to this approach. Liu et al. \cite{liu2024kan} stated not only that an \(\alpha = 4\) was the best scaling exponent but also showed that KANs achieve better performance at lower \(\alpha \) values that MLPs struggle with\cite{liu2024kan}.

While the thought of being able to escape the curse of dimensionality is good. We have to ask the question, how does the quality of input features affect the performance of KANS? In MLPs, there is a direct correlation which can be seen between the quality of input features and the performance. Generally, high informative features lead to performance improvements. But KANS, according to what has been stated above, offers us freedom from the curse of dimensionality allowing us to increase the number of input features. Additionally, Liu et al. \cite{liu2024kan} have shown that KANs can choose intermediate edges by turning them on and off to generate the best approximations through gradient descent \cite{liu2024kan}. In this context, given an arbitrary (both in number and quality) mapping of input features to labeled outputs with no meaningful relation, does a KAN still generate a functional approximation with an error \(\epsilon\) within \(\epsilon \leq CG^{-k-1+m}\) ?\cite{liu2024kan}

\section{Discussion and Conclusion}

Before the advent of deep learning, real-world systems were largely modeled using traditional analytical, statistical, and numerical techniques aimed at defining functional relationships between inputs and outputs or capturing probability distributions of specific mappings. Deep learning has advanced this goal, handling large datasets and uncovering intricate relationships through complex, often opaque neural architectures. Yet, the need for interpretable functional mappings remains, as the complexity of modern deep learning models often surpasses our analytical grasp. Here, the Kolmogorov-Arnold theorem  gains relevance, offering a framework to decompose high-order polynomial functions into compositions of simpler univariate functions. This approach supports interpretable modeling of complex systems. In neural network adaptations, called Kolmogorov-Arnold Networks demonstrate considerable potential for function approximation, opening new applications for deep learning where interpretability and computational efficiency are critical.
KANs are especially effective when the system being modeled can be assumed to follow an underlying polynomial structure. Under this assumption, the Kolmogorov-Arnold theorem ensures that the system’s behavior can be captured as a composition of univariate functions, which KANs are designed to learn. This makes KANs valuable in scenarios where such a polynomial structure is plausible. While it can be challenging to prove this structure in any given system, it is often easier to disprove by identifying discontinuities or irregularities. Thus, in cases where a polynomial form cannot be ruled out, it is reasonable to consider that such a representation may exist, enabling complex polynomial relationships to be broken down into interpretable univariate components. 

The applications of KAT and KANs span a broad spectrum of fields. In any domain where a polynomial representation cannot be ruled out, these theories can be applied to enable advancements in scientific modeling, computer vision, time series analysis, event detection, and even event generation. For example, in image recognition, deep learning models extract features from pixel data and create heuristics that assist in classifying images. This hints at the potential for a functional mapping between the input data or heuristics and image labels, positioning KANs as a possible alternative for image classification tasks. KANs also hold promise in video analysis, where high dimensionality leads to large input feature sets. Each video frame serves as a sizable input vector, and when combined with temporal information, the dimensionality grows rapidly, often resulting in the curse of dimensionality. This high dimensionality likely underpins some of the challenges in state-of-the-art video analysis models. With their demonstrated efficiency in utilizing fewer parameters than traditional multi-layer perceptrons for scientific tasks, KANs could offer distinct advantages in video analysis, especially in multi-spatial detection applications where dimensionality is a critical factor. KANs have shown competitive accuracy sometimes even surpassing MLPs highlighting their potential for tasks that require parameter efficiency and complex approximations.

Thus, KAT and KANs provide foundational theories that form a practical and interpretable framework for neural network-based function approximation and by decomposing complex, high-dimensional behaviors into univariate components, KANs present a promising approach to modeling complex systems, particularly in areas where high-dimensional data and intricate patterns pose challenges for traditional models. As KANs continue to be explored in real-world applications, their parameter efficiency and approximation capabilities suggest significant innovation potential, particularly in high-dimensional tasks like video analysis. This potential paves the way for new advancements in learning model performance and opens doors for further applications in diverse fields.


\bibliographystyle{unsrt}  
\bibliography{references}

\begin{thebibliography}{10}

\bibitem{liu2024kan}
Ziming Liu, Yixuan Wang, Sachin Vaidya, Fabian Ruehle, James Halverson, Marin Solja{\v{c}}i{\'c}, Thomas~Y Hou, and Max Tegmark.
\newblock Kan: Kolmogorov-arnold networks.
\newblock {\em arXiv preprint arXiv:2404.19756}, 2024.

\bibitem{schmidt2018kolmogorov}
Johannes Schmidt-Hieber.
\newblock The kolmogorov--arnold representation theorem revisited.
\newblock {\em Neural Networks}, 2021.

\bibitem{selitskiy2022kolmogorov}
Stanislav Selitskiy.
\newblock Kolmogorov’s gate non-linearity as a step toward much smaller artificial neural networks.
\newblock In {\em Proceedings of the 24th International Conference on Enterprise Information Systems (ICEIS 2022)}, Luton, U.K., 2022. University of Bedfordshire, School of Computer Science and Technology.

\bibitem{vaca2024kolmogorov}
Cristian~J. Vaca-Rubio, Luis Blanco, Roberto Pereira, and M{\`a}rius Caus.
\newblock Kolmogorov-arnold networks (kans) for time series analysis.
\newblock {\em arXiv preprint arXiv:2405.08790}, 2024.
\newblock Submitted on 14 May 2024, last revised 25 Sep 2024 (this version, v2).

\bibitem{genet2024tkan}
Remi Genet and Hugo Inzirillo.
\newblock Tkan: Temporal kolmogorov-arnold networks.
\newblock {\em arXiv preprint arXiv:2405.07344}, 2024.
\newblock Submitted on 12 May 2024, last revised 5 Jun 2024 (this version, v2).

\bibitem{cheon2024efficacy}
Minjong Cheon.
\newblock Demonstrating the efficacy of kolmogorov-arnold networks in vision tasks.
\newblock {\em arXiv preprint arXiv:2406.14916}, 2024.
\newblock Submitted on 21 Jun 2024.

\bibitem{yu2024kan_mlp_comparison}
Runpeng Yu, Weihao Yu, and Xinchao Wang.
\newblock Kan or mlp: A fairer comparison.
\newblock {\em arXiv preprint arXiv:2407.16674}, 2024.
\newblock Accessed: 2024-10-28.

\bibitem{bozorgasl2024wavkan}
Zavareh Bozorgasl and Hao Chen.
\newblock Wav-kan: Wavelet kolmogorov-arnold networks.
\newblock {\em arXiv preprint arXiv:2405.12832}, 2024.
\newblock Submitted on 21 May 2024, last revised 27 May 2024 (this version, v2).

\bibitem{liu2024kan2}
Ziming Liu, Pingchuan Ma, Yixuan Wang, Wojciech Matusik, and Max Tegmark.
\newblock Kan 2.0: Kolmogorov-arnold networks meet science.
\newblock {\em arXiv preprint arXiv:2408.10205}, 2024.
\newblock 27 pages, 14 figures.

\bibitem{koenig2024kan}
Benjamin~C. Koenig, Suyong Kim, and Sili Deng.
\newblock Kan-odes: Kolmogorov–arnold network ordinary differential equations for learning dynamical systems and hidden physics.
\newblock {\em Computer Methods in Applied Mechanics and Engineering}, 432, Part A:117397, 2024.

\bibitem{howard2024finite}
Amanda~A. Howard, Bruno Jacob, Sarah~H. Murphy, Alexander Heinlein, and Panos Stinis.
\newblock Finite basis kolmogorov-arnold networks: Domain decomposition for data-driven and physics-informed problems.
\newblock {\em arXiv preprint arXiv:2406.19662}, 2024.
\newblock Submitted on 28 Jun 2024.

\bibitem{kundu2024kanqas}
Akash Kundu, Aritra Sarkar, and Abhishek Sadhu.
\newblock Kanqas: Kolmogorov-arnold network for quantum architecture search.
\newblock {\em arXiv preprint arXiv:2406.17630}, 2024.
\newblock Submitted on 25 Jun 2024, last revised 22 Jul 2024 (this version, v2).

\bibitem{samadi2024smooth}
Moein~E. Samadi, Younes Müller, and Andreas Schuppert.
\newblock Smooth kolmogorov arnold networks enabling structural knowledge representation.
\newblock {\em arXiv preprint arXiv:2405.11318}, 2024.

\bibitem{hilbert1984mathematical}
David Hilbert.
\newblock {\em Mathematical Problems}.
\newblock Chapman and Hall/CRC, 1st edition, 1984.

\bibitem{hilbert1927mathematische}
David Hilbert, Albert Einstein, Otto Blumenthal, and Constantin Carathéodory, editors.
\newblock {\em Mathematische Annalen}, volume~97.
\newblock Verlag von Julius Springer, Berlin, 1927.
\newblock Founded by Alfred Clebsch and Carl Neumann; continued by Felix Klein, with contributions by Ludwig Bieberbach, Harald Bohr, L. E. J. Brouwer, Richard Courant, Walther v. Dyck, Otto Hölder, Theodor v. Kármán, Arnold Sommerfeld, among others.

\bibitem{wikipedia_kolmogorov_arnold}
{Wikipedia contributors}.
\newblock Kolmogorov–arnold representation theorem, 2024.
\newblock Accessed: 2024-10-28.

\bibitem{galois1962theory}
Évariste Galois.
\newblock {\em The Mathematical Writings of Évariste Galois}.
\newblock Oxford University Press, Oxford, 1962.
\newblock Translated and edited by Peter M. Neumann.

\bibitem{kolmogorov1957representation}
Andrey~N. Kolmogorov.
\newblock On the representation of continuous functions of several variables by superpositions of continuous functions of one variable and addition.
\newblock {\em Doklady Akademii Nauk SSSR}, 114:953--956, 1957.
\newblock Translated in *American Mathematical Society Translations*, Series 2, Vol. 28, pp. 55–59, 1963.

\bibitem{arnold1958representation}
Vladimir~I. Arnold.
\newblock On functions of three variables.
\newblock {\em Doklady Akademii Nauk SSSR}, 114:679--681, 1958.
\newblock Extended Kolmogorov’s theorem to higher dimensions; translated in *American Mathematical Society Translations*, Series 2, Vol. 28, pp. 59–61, 1963.

\bibitem{azam2024suitability}
Basim Azam and Naveed Akhtar.
\newblock Suitability of kans for computer vision: A preliminary investigation.
\newblock {\em School of Computing and Information Systems, The University of Melbourne}, 2024.
\newblock Email: \{basim.azam, naveed.akhtar1\}@unimelb.edu.au.

\bibitem{hou2024comprehensive_kan}
Yuntian Hou, Di~Zhang, Jinheng Wu, and Xiaohang Feng.
\newblock A comprehensive survey on kolmogorov arnold networks (kan).
\newblock {\em arXiv preprint arXiv:2407.11075v1}, 2024.
\newblock Accessed: 2024-10-28.

\bibitem{burden2010numerical}
Richard~L. Burden and J.~Douglas Faires.
\newblock {\em Numerical Analysis}.
\newblock Brooks Cole, Boston, 9th edition, 2010.
\newblock Covers Lagrange and Newton interpolation, spline interpolation, and error estimates.

\bibitem{wikipedia_interpolation}
{Wikipedia contributors}.
\newblock Interpolation, 2024.
\newblock Accessed: 2024-10-28.

\bibitem{wikipedia_polynomial_interpolation}
{Wikipedia contributors}.
\newblock Polynomial interpolation, 2024.
\newblock Accessed: 2024-10-28.

\bibitem{wikipedia_spline_interpolation}
{Wikipedia contributors}.
\newblock Spline interpolation, 2024.
\newblock Accessed: 2024-10-28.

\bibitem{wikipedia_bspline}
{Wikipedia contributors}.
\newblock B-spline, 2024.
\newblock Accessed: 2024-10-28.

\bibitem{deboor2001splines}
Carl de~Boor.
\newblock {\em A Practical Guide to Splines}.
\newblock Springer-Verlag, New York, revised edition edition, 2001.
\newblock Covers B-splines, spline interpolation, and computational aspects.

\bibitem{goodfellow2016deep}
Ian Goodfellow, Yoshua Bengio, and Aaron Courville.
\newblock {\em Deep Learning}.
\newblock MIT Press, 2016.
\newblock Covers perceptron architectures, activation functions, and the theoretical foundations of neural networks.

\bibitem{wikipedia_perceptron}
{Wikipedia contributors}.
\newblock Perceptron, 2024.
\newblock Accessed: 2024-10-28.

\bibitem{wikipedia_universal_approximation_theorem}
{Wikipedia contributors}.
\newblock Universal approximation theorem, 2024.
\newblock Accessed: 2024-10-28.

\bibitem{cybenko1989approximation}
George Cybenko.
\newblock Approximation by superpositions of a sigmoidal function.
\newblock {\em Mathematics of Control, Signals and Systems}, 2(4):303--314, 1989.

\bibitem{sharma2020neural_scaling}
Utkarsh Sharma and Jared Kaplan.
\newblock A neural scaling law from the dimension of the data manifold.
\newblock {\em arXiv preprint arXiv:2004.10802}, 2020.
\newblock Accessed: 2024-10-28.

\bibitem{barron1993approximation}
Andrew~R. Barron.
\newblock Universal approximation bounds for superpositions of a sigmoidal function.
\newblock {\em IEEE Transactions on Information Theory}, 39(3):930--945, 1993.

\end{thebibliography}

\end{document}